\documentclass[10pt,twocolumn,letterpaper]{article}

\usepackage{cvpr}              

\usepackage{paralist}
\usepackage[ruled, vlined, linesnumbered]{algorithm2e}
\usepackage[noend]{algpseudocode}
\usepackage{booktabs}
\usepackage{multirow}
\usepackage{amsmath}
\usepackage{amsthm}
\usepackage{amssymb}
\usepackage{newfloat}
\usepackage{listings}
\usepackage{graphicx}
\usepackage{booktabs}
\usepackage{makecell}
\usepackage{bbm}

\usepackage[pagebackref,breaklinks,colorlinks]{hyperref}

\usepackage[capitalize]{cleveref}
\crefname{section}{Sec.}{Secs.}
\Crefname{section}{Section}{Sections}
\Crefname{table}{Table}{Tables}
\crefname{table}{Tab.}{Tabs.}

\newcommand{\Approach}[1]{Sibling-Attack}

\newcommand{\bj}[1]{\textcolor{orange}{(bj: #1})}

\def\cvprPaperID{754} 
\def\confName{CVPR}
\def\confYear{2023}

\begin{document}

\title{Sibling-Attack: Rethinking Transferable Adversarial Attacks against Face Recognition}


\author{Zexin Li$^{1}$\footnotemark[1]
\xspace\xspace\xspace    Bangjie Yin$^{3}$\footnotemark[1]
\xspace\xspace\xspace    Taiping Yao$^{3}$
\xspace\xspace\xspace    Junfeng Guo$^{2}$
\xspace\xspace\xspace    Shouhong Ding$^{3}$\footnotemark[2]
\xspace\xspace\xspace    Simin Chen$^{2}$
\xspace\xspace\xspace    Cong Liu$^{1}$\footnotemark[2]\\
$^{1}$University of California, Riverside\\
$^{2}$The University of Texas at Dallas\\
$^{3}$Tencent\\
{\tt\small \{zli536, congl\}@ucr.edu, \{junfeng.guo, simin.chen\}@utdallas.edu,} \\
{\tt\small \{bangjieyin, taipingyao, ericshding\}@tencent.com} 
}
\maketitle

\renewcommand\arraystretch{1.1} 
\renewcommand{\thefootnote}%
{\fnsymbol{footnote}}
\footnotetext[1]{indicates equal contributions.} 
\footnotetext[2]{indicates corresponding author.}

\begin{abstract}
A hard challenge in developing practical face recognition (FR) attacks is due to the black-box nature of the target FR model, i.e., inaccessible gradient and parameter information to attackers. While recent research took an important step towards attacking black-box FR models through leveraging transferability, their performance is still limited, especially against online commercial FR systems that can be pessimistic (e.g., a less than 50\% ASR--attack success rate on average). Motivated by this, we present \emph{\Approach{}}, a new FR attack technique for the first time explores a novel \textit{multi-task} perspective (i.e., leveraging extra information from multi-correlated tasks to boost attacking transferability). Intuitively, \emph{\Approach{}} selects a set of tasks correlated with FR and picks the Attribute Recognition (AR) task as the task used in \emph{\Approach{}} based on theoretical and quantitative analysis. \emph{\Approach{}} then develops an optimization framework that fuses adversarial gradient information through (1) constraining the cross-task features to be under the same space, (2) a joint-task meta optimization framework that enhances the gradient compatibility among tasks, and (3) a cross-task gradient stabilization method which mitigates the oscillation effect during attacking. Extensive experiments demonstrate that \emph{\Approach{}}  outperforms state-of-the-art FR attack techniques by a non-trivial margin, boosting ASR by 12.61\% and 55.77\% on average on state-of-the-art pre-trained FR models and two well-known, widely used commercial FR systems.

\end{abstract}

\section{Introduction}
\label{sec:intro}

\begin{figure}[!t]
    \centering
    \includegraphics[width=0.4\textwidth]{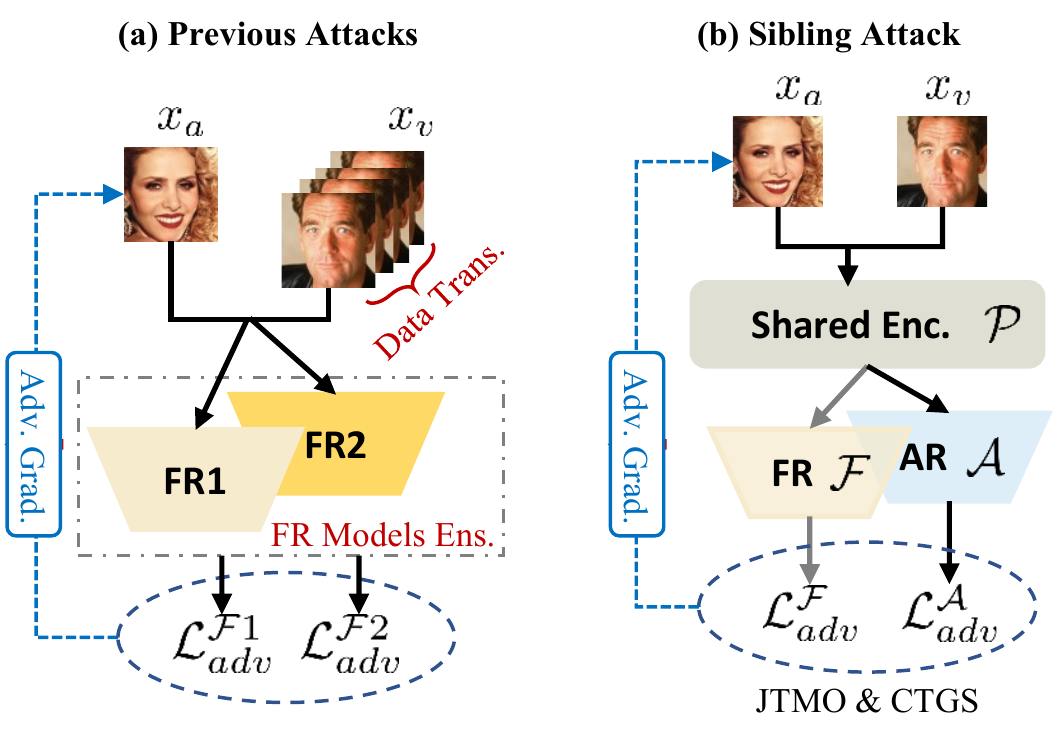}
    \caption{
     Under the single task, previous attacks \textbf{(a)} boost transferability by attacking multiple models or using various sampling or augmentation strategies. Nevertheless, in the proposed \Approach{} \textbf{(b)}, we adopt the Attribute Recognition (AR) as the auxiliary task to improve the transferability. And we utilize the hard-parameter sharing architecture from~\cite{caruana1997multitask} as the attacking backbone. 
    }
    
    \label{fig:architecture}
\end{figure}

Deep Neural Networks (DNNs) have demonstrated significant success in various applications, especially for face recognition~\cite{deng2019arcface, schroff2015facenet}. Despite these achievements, recent research has revealed that DNN-based face recognition (FR) models may be susceptible to adversarial attacks~\cite{szegedy2013intriguing, goodfellow2014explaining, carlini2017towards}. In practical attacking scenarios, the victim FR model's parameters are inaccessible to the attackers~\cite{dong2019efficient, chen2017zoo, papernot2017practical, wang2021delving, liu2016delving}, i.e., the attacker has to perform attacks under black-box settings. One feasible black-box attacking strategy is to craft transferable adversarial examples by attacking a white-box surrogate model. On the face recognition task, recent research (e.g., optimization-based methods~\cite{dong2018boosting,lin2019nesterov,wang2021enhancing}, model-ensemble training~\cite{dong2018boosting,liu2016delving} and input data transformations~\cite{dong2019evading,xie2019improving,wu2021improving}) has shown efficacy on boosting the attacking transferability. Essentially, those methods prevent the adversarial examples from over-fitting to a single model/image by fusing auxiliary gradient information from ensemble models or various sampling/augmenting strategies.
However, their performance against online commercial FR systems can be rather pessimistic (e.g., a less than 50\% attack success rate on average as shown in our evaluation).

Motivated by this, we obtain an important insight by understanding such pessimism is that existing methods collect adversarial gradients only from the single task and thus overlook the potential possibilities to further improve transferability, as illustrated in Fig.~\ref{fig:architecture}\textbf{(a)}.
Recent multi-task learning (MTL) methods~\cite{caruana1997multitask, liu2019end, zhou2020pattern, standley2020tasks} have indicated that the multi-task or joint-task training among the correlated tasks can learn more robust and general features and thus improve the overall generalizability.
Inspired by this, we seek to improve the FR task's attacking transferability within the cross-task scope. 
To explore the FR attacking transferability under a multi-task setting, there are two challenges: 1) identifying an appropriate auxiliary task as a suitable candidate for FR task when performing multi-task attacks, and 2) how to fully utilize the adversarial information from two tasks thus boosting transferability.

We assume that a face-related task, which can provide relevant but diverse adversarial gradients information to complement the inherently absent adversarial knowledge for the target FR task, could be deemed as a good auxiliary task candidate, named \emph{sibling task}. The empirical observations of previous works\cite{taherkhani2018deep,diniz2020face} have proved that the AR model can learn robust identity features, which can be used to enhance the FR's recognition robustness. Also, in turn, FR features implicitly encode latent facial attribute features. In addition, we conduct quantitative results to
show the effectiveness of the AR task. To this end, we leverage a correlated AR task as the sibling task to improve the attacking transferability, i.e., \emph{\Approach{}}.

Since big variance exists in the feature and gradient spaces of different tasks\cite{mao2020multitask,sener2018multi,dong2022neural}, direct optimization over FR and AR models will lead to a limited attacking transferability without considering the better gradients fusion and stabilized training strategies. To address the issues, in \emph{\Approach{}}, we first adopt the hard-parameter sharing architecture derived from~\cite{caruana1997multitask} as our backbone attacking framework to constrain them within the same feature space, as shown in Fig.~\ref{fig:architecture}\textbf{(b)}. Next, we design an alternating joint-task meta optimization (JTMO) algorithm based on the high-level spirit of meta-learning~\cite{MAML,nichol2018reptile,shao2020regularized} to further improve the gradient compatibility between two tasks. Finally, to mitigate the training oscillation effect, we propose a cross-task gradient stabilization (CTGS) strategy for stabilizing the adversarial example optimization.


Extensive experiments demonstrate that \Approach{}  outperforms state-of-the-art FR attack techniques by a non-trivial margin, boosting the attack success rate by 12.61\% and 55.77\% on average on state-of-the-art pre-trained FR models and two well-known, widely used commercial FR systems, Face++ face recognition~\cite{MEGVII} and Microsoft face API~\cite{Microsoft}. 
Notably, \Approach{} yields 86.50\% and 96.10\% ASR on attacking the widely used Face++ commercial face API on two common datasets, while the state-of-the-art only reaches 58.10\% and 64.30\%, respectively.


We summarize our contributions as: \textbf{1)} We propose to generate highly transferable adversarial examples against face recognition by utilizing the adversarial information from the related AR task. \textbf{2)} We propose a novel \emph{\Approach{}} method which jointly learns the adversarial information from multiple tasks in a more effective manner. \textbf{3)} Evidenced by extensive experiments, the ASR of \emph{\Approach{}} significantly outperforms current SOTA single-task attacks on the widely-adopted and large-scale FR benchmarks, particularly, several \textit{online commercial FR systems}, which is aligned with our assumptions and analyses.

\begin{figure*}[!t]
    \centering
    \includegraphics[width=0.85\textwidth]{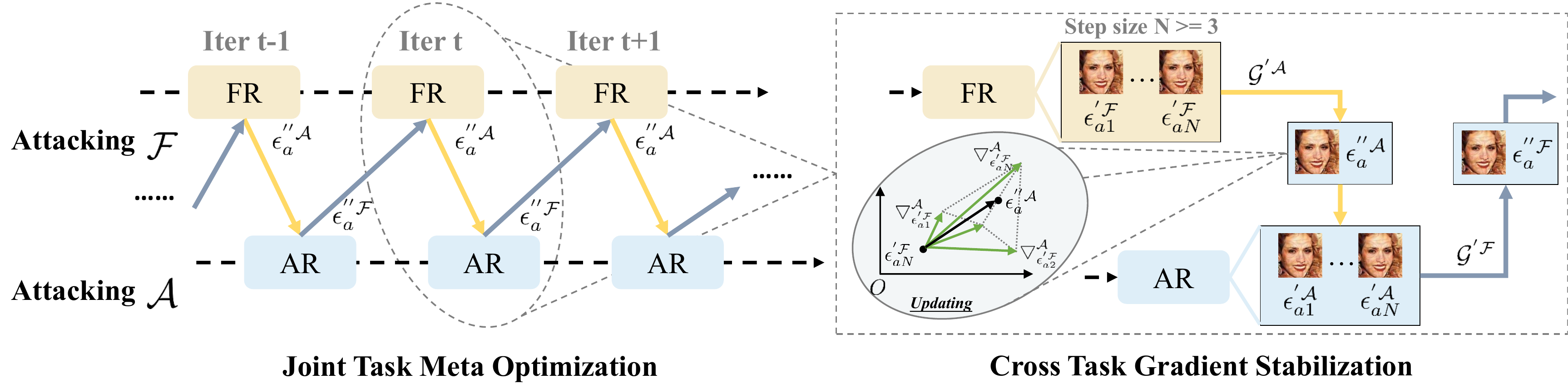}
    \caption{The optimizing process of \emph{\Approach{}}. The first row illustrates Joint-Task Meta Optimization (\textit{JTMO}) and the second row exhibits Cross-Task Gradient Stabilization (\textit{CTGS}). JTMO alternatively selects models from different tasks for each iteration (Sec.~\ref{sec: JTMO}). CTGS stabilizes the cross-task gradient via historical information (Sec.~\ref{sec: CTGS}). }
    \label{fig:architecture_2}
\end{figure*}

\section{Related Work}

\subsection{Adversarial Attacks}

Adversarial attacks raise significant concern in machine learning due to their potential impact on security and safety-critical applications.~\cite{xie2019improving,dong2018boosting,madry2017towards,kurakin2016adversarial,nnreverse,chen2022nicgslowdown, chen2022nmtsloth, chen2022deepperform, aeva,guo2023scale,guo2020practical,deepbill,kong2020physgan}
Recently, several approaches have been proposed to enhance the transferability of adversarial attacks by designing underlying optimization algorithms based on the BIM~\cite{kurakin2016adversarial} or PGD~\cite{madry2017towards}. For instance, MI-FGSM~\cite{dong2018boosting} incorporates momentum to BIM and uses ensemble models to craft adversarial samples. VMI-FGSM~\cite{wang2021enhancing} alleviates the gradient variance to boost the performance. TAP~\cite{zhou2018transferable} shows that attacking intermediate feature maps could help to generate more transferable adversarial examples. DI-FGSM~\cite{xie2019improving} proposes a method to increase the diversity of the inputs by randomly altering the input data. Wu. \textit{et al.}~\cite{wu2021improving} makes adversarial examples insensitive to distortions by leveraging a transformation network. Xiong \textit{et al.}~\cite{xiong2022stochastic} focus on reducing stochastic variance to boost ensemble transferable attacking performance. 
NAA~\cite{zhang2022improving} improves the performance of transferable attacks on the feature level by more accurate neuron importance estimations. TAIG~\cite{huang2022transferable} boosts transferability by 
optimizing standard objective functions, 
exploiting attention maps, and smoothing decision surfaces.

Regarding the transferable digital adversarial attacks against the FR task, Adv-Face~\cite{deb2020advfaces} employs a GAN-based framework to address the over-fitting problem. DFANet~\cite{zhong2020towards}  applies dropout layers to boost attacking transferability. On the other hand, a set of work studies transferable physical attacks against FR systems using patch-based methods. Adv-Glasses~\cite{sharif2016accessorize} and Adv-Hat~\cite{komkov2019advhat} perform physical adversarial attacks by injecting patched hats or eyeglasses. The most recent work~\cite{yin2021adv,jia2022adv}, generates imperceptible perturbations of specific makeup and facial attributes. Unlike previous work boosting transferability by performing a single task white-box attack,
we propose a new framework to craft transferable attacks against the FR model by leveraging the AR task's information. 

\subsection{Multi-task Learning}
Multi-task Learning (MTL)~\cite{caruana1997multitask,uninet,guo2020multi,wang2017multi,taherkhani2018deep} is to learn multiple tasks simultaneously to improve the accuracy of each task compared with single-task learning (STL)~\cite{miao2021data,chen-etal-2021-revisiting,chen-etal-2022-generate, nnreverse, chen2020denas}. 
Several existing works have proved the strong correlations between FR and AR tasks. Diniz \textit{et al.}\cite{diniz2020face} illustrates that the FR model implicitly encodes latent attribute features in the representations, and the hidden layer of the FR model can be used to perform attribute prediction. Hu \textit{et al.}\cite{hu2017attribute} claims that models for the AR task can learn more robust features and thus can be used to improve FR robustness. Taherkhani \textit{et al.} \cite{taherkhani2018deep} leverage AR models as a soft modality to enhance the performance of FR models. Wang \textit{et al.}\cite{wang2017multi} utilize a multi-task framework to boost training performance on both FR and AR tasks. 
Ghamizi et al.~\cite{ghamizi2022adversarial} and Mao et al.~\cite{mao2020multitask} have claimed that multi-task training can learn more adversarial robust features.

Recent concurrent work has studied adversarial attacks against multi-task models.  MTA~\cite{guo2020multi} perform white-box attacks adversarial attacks against hard parameter sharing architecture multi-task learning models. UniNet~\cite{uninet} introduces adversarial attacks to better explore the relationship between multi-tasks in an autonomous driving scenario.
Nevertheless, there exist significant differences from \Approach{}: 1) \emph{\Approach{}} focuses on improving black-box attacking transferability rather than maintaining the efficacy of white-box attacks; 2) \emph{\Approach{}} proposes JTMO and CTGS optimization strategies to further boost transferability (in Sec.~\ref{sec:method}). 3) \emph{\Approach{}} evaluates transferable attacks against online commercial platforms and significantly improves the performance.


\section{Methodology}
\label{sec:method}
\subsection{Overview}
\label{sec:overview}

The targeted adversarial attack against FR, i.e., \textit{impersonation attack} \cite{deb2020advfaces}, spoofs the target FR model to misidentify the attacker as the same identity as the target, which is more challenging and malicious than \textit{dodging attack} \cite{deb2020advfaces} in the real world. Therefore, this paper mainly focuses on the impersonation attack as in \cite{yin2021adv,zhong2020towards}. The objective of the impersonation attack can be formulated as follows:

\begin{equation}
\min_{\epsilon_a}\ \mathcal{L}(x_a+\epsilon_a, x_v),\ s.t.\ \| \epsilon_a \|_{p} \leq \xi
\label{eq: attacks}
\end{equation}
where $ x_a \in  \mathcal{R}^{H \cdot W \cdot C}  $  is the attacking face and $x_v \in  \mathcal{R}^{H \cdot W \cdot C}$ is the target victim face. The perturbation $\epsilon_a \in [0, 1]^{H \cdot W \cdot C}$ to the attacker is constrained by the $\ell_p$-norm ($p \in \{0,2,\infty\}$). In this work, we use $\ell_\infty$-norm as the metric following~\cite{dong2018boosting,madry2017towards,wang2021enhancing,zhou2018transferable}. $ \xi $ is a small constant to bound $ \epsilon_a $. $ \mathcal{L}(\cdot) $ denotes the adversarial loss function. 

\subsection{\Approach{} Framework}

As shown in Fig.~\ref{fig:architecture}\textbf{(b)}, we adopt a prevelant hard parameter sharing architecture~\cite{baxter1997bayesian,caruana1997multitask} as the backbone in \emph{\emph{\Approach{}}} to avoid large feature variance~\cite{mao2020multitask}. Our white-box surrogate model shown in Fig.~\ref{fig:architecture}\textbf{(b)} is denoted as $\mathcal{S}(\mathcal{P}; \mathcal{F}; \mathcal{A})$, with a sharing-parameter encoder $ \mathcal{P} $ as its first component. Then the surrogate model branches off into two sub-networks: an FR branch $ \mathcal{F} $, and an AR branch $ \mathcal{A} $. Given an attacking image $ x_a  $ and a target image $ x_v $, our goal is to generate adversarial examples $ x_{adv} $ through $ \mathcal{S} $ to fool the black-box target FR model $ \mathcal{T} $. Specifically, for each $ x_a $ and $ x_v $, each branch of $\mathcal{S}$ compute their corresponding output high-level feature vectors $ \{f^{\mathcal{F}}_a ,  f^{\mathcal{F}}_v\} $ through $ \mathcal{F} $ and $ \{f^{\mathcal{A}}_a ,  f^{\mathcal{A}}_v\}$ through $ \mathcal{A} $, respectively. These features are then used to compute the corresponding adversarial loss for targeted attacks against FR as follows:

\begin{equation}
\mathcal{L}_{adv}^{*}=1-cos(f_a^{*}, f_v^{*})
\label{eq: cos_simi}
\end{equation}
where $ * \in \left\{ \mathcal{F}, \mathcal{A} \right\} $ and we use the cosine value \cite{deb2020advfaces,yin2021adv,zhong2020towards} between two feature vectors as the evaluation metric to measure their similarity. Based on that, the main objective of the joint impersonation attack is designed as follows: 

\begin{equation}
\min_{\epsilon_a}\ \lambda_1 \cdot \mathcal{L}_{adv}^{\mathcal{F}} + \lambda_2 \cdot \mathcal{L}_{adv}^{\mathcal{A}},\ s.t.\ \| \epsilon_a \|_{p} \leq \xi
\label{eq: intuitve_atack}
\end{equation}
where $ \lambda_1 $ and $ \lambda_2 $ are the trade-off hyper-parameters. 


\subsection{Joint-Task Meta Optimization} 
\label{sec: JTMO}

Revisiting the existing meta-learning frameworks, 
several researchers ~\cite{nichol2018reptile,finn2017model,shao2020regularized} have proven that alternatively adopting gradients can improve the cross-dataset compatibility of conducting feature learning, thus enhancing generalizability. This fact motivates us to craft transferable adversarial examples by obtaining better gradient compatibility between two tasks.
Therefore, we propose a new optimization strategy targeting adversarial scenarios, namely Joint-Task Meta Optimization \textit{(JTMO}). As shown in Fig.~\ref{fig:architecture_2},
in \textit{JTMO}, we imitate the parameter updating strategy of meta-learning instead of directly calculating weighted average adversarial losses for two tasks.

To generate the adversarial examples, we have to iteratively modify the pixels in $ x_a $ by adding a perturbation $ \epsilon_a$. For each iteration, we alternately choose one branch from $ \mathcal{S} $, and then perform forward- and back-propagation to calculate the gradients from the corresponding adversarial losses, $ \mathcal{L}_{adv}^{\mathcal{F}} $ or $ \mathcal{L}_{adv}^{\mathcal{A}} $. The order of branch selection won't affect the final performance. For each branch in each iteration, the updated perturbation $\epsilon_a^{'}$ can be computed by:

\begin{equation}
\epsilon_a^{'} \leftarrow \Pi \left\{ \epsilon_a - \alpha \cdot sign (\gamma_1 \cdot \bigtriangledown_{\epsilon_a} \mathcal{L}_{adv}^{*}(x_a + \epsilon_a, x_v)) \right \}
\label{eq: metatrain}
\end{equation}
where $ \Pi \left \{ \cdot \right \} $ denotes the projection function ensured by $ \ell_\infty $ constrain. $ \alpha $ is learning rate, $ \gamma_1 $ is the updating hyper-parameter, and also $ * \in \left\{ \mathcal{F}, \mathcal{A} \right\} $. Then, we utilize the updated perturbation $\epsilon_a^{'}$ to compute the adversarial losses for the remaining un-chosen branch in the  $\mathcal{S}$. Thus, we compute $ \mathcal{L}_{adv}^{*'} $ based on $ \epsilon_a^{'} $. Finally, we aggregate all the gradient information to update the perturbation as follows:

\vspace{-2mm}
\begin{equation}
\begin{split}
\epsilon_a^{''} \leftarrow \Pi \left \{\epsilon_a^{'} - \alpha \cdot sign (\gamma_2 \cdot \bigtriangledown_{\epsilon_a^{'}} \mathcal{L}_{adv}^{*'}(x_a^{'} + \epsilon_a^{'}, x_v)) \right \}
\end{split}
\label{eq: metaopt}
\end{equation}
where $x_a^{'} = x_a + \epsilon_a$, $ \gamma_2 $ is the updating hyper-parameter and $ \epsilon_a^{''} $ is the output of adversarial perturbations for each iteration. 
Inspired by meta-learning, our optimization strategy first 
collect gradients alternatively from two branches w.r.t the perturbation parameters, then adopt the gradients to optimize $ \epsilon_a^{''} $ alternately between two tasks for every iteration to obtain optimization compatibility.

\begin{algorithm}[t!]
\caption{The proposed attacking method}
\label{alg:algorithm}
{\bf Require:}
Attacking images $ x_a \in \mathcal{R}^{H \cdot W \cdot C}$; vcitim images $ x_v \in \mathcal{R}^{H \cdot W \cdot C}$; adversarial perturbations $ \epsilon_a \in [0, 1]^{H \cdot W \cdot C}$; pre-trained multi-task model $ \mathcal{S}(\mathcal{P}; \mathcal{F}; \mathcal{A}) $; iterations $ T $; updating step size $ N $. \\
{\bf Initialization:} 
Adversarial example parameters $ \epsilon_a $;hyperparameters $\gamma_1 $, $ \gamma_2 $, $ \gamma_3 $; learning rate $ \alpha $.\\
{\bf Ensure:}
Perturbation parameters $ \epsilon_a^{opt} $.\\
$ x_{adv} = x_a $;\\
\For{each $ t\in T $}
{
    Alternatively select one task branch, such as $ \mathcal{F} $;\\
    Update $ \mathcal{E}^{'\mathcal{F}} = \left \{ \emptyset \right \} $;\\
    \For{each $ i \in N $}
    {
        Calculate $ \mathcal{L}_{adv}^{\mathcal{F}} $ on ($ x_{adv} $, $ x_v $) with Eq.~\ref{eq: cos_simi};\\
        Obtain $ \epsilon_{ai}^{'\mathcal{F}} $ by $ \mathcal{L}_{adv}^{\mathcal{F}} $ with Eq.~\ref{eq: metatrain};\\
        Append $ \epsilon_{ai}^{'\mathcal{F}} $ to $ \mathcal{E}^{'\mathcal{F}} $;\\
        Update $ x_{adv} = x_{adv} + \epsilon_{ai}^{'\mathcal{F}} $;
    }

    Obtain $ \mathcal{G}^{'\mathcal{A}} = \left \{ \bigtriangledown_{\epsilon_{a1}^{'\mathcal{F}}}^{\mathcal{A}}, ..., \bigtriangledown_{\epsilon_{aN}^{'\mathcal{F}}}^{\mathcal{A}} \right \}$ from another branch $ \mathcal{A} $ based on $ \mathcal{E}^{'\mathcal{F}} $;\\
    Update $ \epsilon_{a}^{''\mathcal{A}} $ with Eq. \ref{eq: emetaopt};\\
    Update $ x_{adv} = x_{adv} + \epsilon_{a}^{''\mathcal{A}} $;\\
    
}
\label{endfor}
$ \epsilon_a^{opt} =\epsilon_{a}^{''\mathcal{A}}$; \\
\Return{$ \epsilon_a^{opt}$}
\end{algorithm}

\subsection{Cross-Task Gradient Stabilization}
\label{sec: CTGS}




Updating adversarial perturbations across two tasks may inevitably cause a side-effect of oscillation and lead to a sub-optimal solution. This side-effect can be attributed to the fact that the two different tasks have different gradient updating directions\cite{sener2018multi}. Recent methods of single-task adversarial attacks\cite{dong2018boosting,wang2021enhancing} have claimed that historical gradients and appropriate gradients aggregation could stabilize the optimizing process, thus boosting attacking transferability. Inspired by them, we design a new updating strategy, namely Cross-Task Gradient Stabilizing(\textit{CTGS}), to further improve the attacking transferability of \emph{\Approach{}}.

As shown in Fig.~\ref{fig:architecture_2}, at each iteration of the optimizing process, we define an updating step size $ N $ for the selected task branch, e.g., $ \mathcal{F} $. Then $ N $ adversarial perturbations, $ \mathcal{E}^{'\mathcal{F}} = \left\{ \epsilon_{a1}^{'\mathcal{F}}, ..., \epsilon_{aN}^{'\mathcal{F}} \right\} $ , can be crafted iteratively by consecutive steps updating with Eq.~\ref{eq: metatrain} based on $ \mathcal{F} $. Next, we add the perturbations to the attacking image $ x_a $ to generate the adversarial examples and send them into another task branch $ \mathcal{A} $ and compute their corresponding gradient maps, $ \mathcal{G}^{'\mathcal{A}} = \left \{ \bigtriangledown_{\epsilon_{a1}^{'\mathcal{F}}}^{\mathcal{A}}, ..., \bigtriangledown_{\epsilon_{aN}^{'\mathcal{F}}}^{\mathcal{A}} \right \}$. Hence, when updating the $ \epsilon_a^{''} $ on the $ \mathcal{A} $, we can derive Eq.~\ref{eq: metaopt} as:

\vspace{-4mm}
\begin{equation}
\epsilon_{a}^{''\mathcal{A}} \leftarrow \Pi \left \{\epsilon_{aN}^{'\mathcal{F}} - \alpha * sign[ \gamma_2 * ( \bigtriangledown_{\epsilon_{aN}^{'\mathcal{F}}}^{\mathcal{A}} + \gamma_3 \sum_{i=1}^{N-1} \bigtriangledown_{\epsilon_{ai}^{'\mathcal{F}}}^{\mathcal{A}})] \right \}
\label{eq: emetaopt}
\end{equation}

Following this updating procedure, the calculated gradients on $\mathcal{A}$ for the historical 
adversarial gradients from $ \mathcal{F} $ are aggregated for stabilizing the current optimization. 
$ \gamma_3 $ is a hyper-parameter to balance the training weights. We choose $ \gamma_3 $ as a small number since the historical adversarial gradients merely provide the auxiliary gradients information rather than dominate the main updating direction. Our strategy enhances the optimizing stability and promotes transferability by utilizing the cross-task gradients of the historical $ N-1 $ adversarial examples from another task branch. The overall procedure of \emph{\Approach{}} is shown in Alg.~\ref{alg:algorithm}.

\begin{table}[t!]
\begin{centering}
\setlength{\tabcolsep}{1.6mm}{
\hspace{-0.1in}\scalebox{0.9}{ %
\begin{tabular}{c}
\begin{tabular}{c|cc|cc}
\hline 
Dataset & \multicolumn{2}{c|}{CelebA-HQ} & \multicolumn{2}{c}{LFW} \tabularnewline
\cline{1-5}
Target Model & IR50 & ResNet101 & IR50 & ResNet101 \tabularnewline
\hline
FR+FR & 73.40 & 76.00 & \textcolor{blue}{75.80} & 78.20  \tabularnewline
FR+FLD & \textcolor{blue}{75.20} & 78.10 & 52.00 & 78.60 \tabularnewline
FR+FP & 66.50 & \textcolor{blue}{85.10} & 71.80 & \textcolor{blue}{83.40} \tabularnewline
\hline
FR+AR(Ours) & \textbf{93.00} & \textbf{93.40} & \textbf{97.60} & \textbf{96.80} \tabularnewline
\hline
\end{tabular}\tabularnewline
\end{tabular}}}
\par\end{centering}
\caption{ASR results for black-box impersonation attacks against different task combinations. Best attack performance results are shown in bold. The $2^{nd}$ place performance is shown in blue.
\label{Tab:FR Related Tasks}}
\end{table}

\section{Experiments}

\subsection{Experimental Setup}
\noindent\textbf{Datasets.} To evaluate the attacking transferability of the proposed \emph{\Approach{}}, we choose two popular face datasets:
\begin{inparaenum}[1)]
     \item \textit{CelebA-HQ}~\cite{karras2017progressive}: The CelebA-HQ dataset is a high-quality update for the CelebA dataset~\cite{liu2015deep}, which consists of 30,000 best-looking facial images. 
     \item \textit{LFW}~\cite{huang2008labeled}: Labeled Faces in the Wild (LFW) is a dataset for face recognition that contains 13,233 images collected on the web of 5,749 different subjects. 
\end{inparaenum}
 We randomly sample 1,000 pairs of different-identity faces for each dataset to evaluate \emph{\Approach{}}'s attacking performance.

\noindent\textbf{Evaluation Metrics.} Following prior works~\cite{deb2020advfaces,yin2021adv,zhong2020towards,li2}, we adopt Attack Success Rate (ASR) for impersonation attack to evaluate \emph{\Approach{}}, which is computed through:
\begin{equation}
    \text{ASR} = \frac{\text{No. of Comparisons} \geq \tau}{\text{Total No. of Comparisons}}
    \label{eq: ASR}
\end{equation}



\noindent Whether an adversarial attack is successful is defined by  the numerator of Eq.~\ref{eq: ASR}, which accepts the similarity scores between the adversarial examples and benign examples from the black box model over the corresponding threshold $\tau$. 

\noindent\textbf{Baselines.}
We compare \emph{\Approach{}} with ten start-of-the-art adversarial attacks, namely, face-based and transfer-based attacks:
\begin{inparaenum}[1)]
    \item \textit{face-based attacks:} Adv-Hat~\cite{komkov2019advhat}, Adv-Glasses~\cite{sharif2016accessorize}, 
    Adv-Face~\cite{deb2020advfaces}, Adv-Makeup~\cite{yin2021adv} and GenAP~\cite{xiao2021improving}.
    \item \textit{transfer-based attacks:} PGD~\cite{madry2017towards}, TAP~\cite{zhou2018transferable}, MI-FGSM~\cite{dong2018boosting}, VMI-FGSM~\cite{wang2021enhancing}.
\end{inparaenum}

\noindent\textbf{Target Model.} Similar to the evaluation in prior works~\cite{yin2021adv}, we choose a mix of the various offline and online commercial FR models to evaluate the transferability of the adversarial examples generated by \emph{\Approach{}}. Specifically, we choose:
\begin{inparaenum}[1)]
    \item \textit{Offline models:} five famous face recognition models: IR152~\cite{deng2019arcface}, IRSE50~\cite{deng2019arcface}, FaceNet~\cite{schroff2015facenet}, IR50~\cite{deng2019arcface}, ResNet101~\cite{he2016deep}. 
    \item \textit{Online models:} two widely used online commercial face recognition systems: Face++~\cite{MEGVII} and Microsoft~\cite{Microsoft}.
\end{inparaenum}
For the offline FR models, we use IR152, IRSE50, and FaceNet as white-box models to generate adversarial examples and evaluate attacking transferability on the other models. 
All the thresholds of offline models are obtained from the images in the LFW dataset~\cite{deb2020advfaces}. 
We set $\tau$ to  (0.277, 0.200) following~\cite{madry2017towards,yin2021adv,zhang2019theoretically} for (IR50, ResNet101). 
For the online FR models, as per the suggestions of platforms, we set $\tau$ as the cosine similarity score at 0.001 FPR (False Positive Rate) level for Face++. For Microsoft, We use the reported query results as the number of successful attacks since Microsoft does not offer the cosine similarity score for different FPR levels and only gives a cosine similarity score and decision result for each query.

\noindent\textbf{AR Model.} For AR models, we use IR152~\cite{deng2019arcface} and Mobileface~\cite{chen2018mobilefacenets} as the backbone networks and train them on MS-Celeb-1M \cite{guo2016ms}, and CelebA-HQ \cite{karras2017progressive}, to guarantee their performance on the AR task. We include the detailed training scheme in the supplementary files due to the page limits.

\noindent\textbf{Implementation Details.} In \emph{\Approach{}}, the structure of the white-box surrogate model is IR152 for both the FR task and the AR task. Following the experimental configuration of previous work~\cite{xiao2021improving}, we set $\xi$ to 40/255 as the $\ell_\infty$ bound as~\cite{dong2018boosting,madry2017towards,wang2021enhancing,zhou2018transferable,li1} for ours and baselines. Meanwhile, the step size $\alpha$ is set to 2/255 while the iteration number $T$ is set to 200 to ensure attack efficacy, and updating step size $N$ is 4. Moreover, we initialize ($\gamma_1, \gamma_2, \gamma_3$) as (0.1, 0.9, 0.01). All competitors strictly adopt their original setting. 
Since our method performs attacks across two different
models, FR and AR, and existing works~\cite{dong2018boosting,liu2016delving} have evidenced the merits of ensemble attacking, we attack two FR models for other competitors to ensure comparison fairness.

\begin{table*}[t!]
\begin{centering}
\setlength{\tabcolsep}{1.6mm}{
\hspace{-0.1in}\scalebox{0.9}{ %
\begin{tabular}{c}
\begin{tabular}{c|c|cc|cc|cc|cc}
\hline 
\multirow{4}{*}{Methods } & Dataset & \multicolumn{8}{c}{CelebA-HQ} \tabularnewline
\cline{2-10}
& Source Model & \multicolumn{4}{c|}{IR152+FaceNet} & \multicolumn{4}{c}{IR152+IRSE50} \tabularnewline
\cline{2-10}
& \multirow{2}{*}{Target Model} & \multicolumn{2}{c|}{Offline Model} & \multicolumn{2}{c|}{Online Model} & \multicolumn{2}{c|}{Offline Model} & \multicolumn{2}{c}{Online Model} \tabularnewline
\cline{3-10}
& & IR50 & ResNet101 & Face++ & Microsoft & IR50 & ResNet101 & Face++ & Microsoft \tabularnewline
\hline
\multirow{6}{*}{Face-based} & Adv-Hat~\cite{komkov2019advhat}
& 1.50 & 6.50 & 1.00 & 0.00 & 3.80 & 8.70 & 0.90 & 0.00 \tabularnewline
& Adv-Glasses~\cite{sharif2016accessorize}
& 0.60 & 8.50 & 3.40 & 0.00 & 5.90 & 9.70 & 4.20 & 0.10 \tabularnewline
& Adv-Face~\cite{deb2020advfaces}
& 58.80 & 64.60 & {\textcolor{blue}{54.90}} & 8.70 & 68.00 & 71.40 & 48.00 & 8.70 \tabularnewline
& Adv-Makeup~\cite{yin2021adv}
& 8.30 & 21.20 & 5.30 & 0.00 & 13.00 & 26.00 & 4.90 & 0.10 \tabularnewline
& GenAP~\cite{xiao2021improving}
& 52.80 & 49.10 & 54.40 & 6.40 & 47.10 & 48.40 & 47.20 & 5.80 \tabularnewline
\hline
\multirow{4}{*}{Transfer-based} & PGD~\cite{madry2017towards}
& 73.40 & 76.00 & 37.20 & 13.00 & {\textcolor{blue}{92.00}} & {\textcolor{blue}{90.80}} & {\textcolor{blue}{58.10}} & 28.70 \tabularnewline
& TAP~\cite{zhou2018transferable}
& 72.80 & 76.20 & 42.90 & {\textcolor{blue}{20.40}} & 88.30 & 87.60 & 52.90 & {\textcolor{blue}{28.90}} \tabularnewline
& MI-FGSM~\cite{dong2018boosting}
& 66.60 & 73.30 & 36.10 & 14.80 & 86.20 & 90.10 & 57.80 & {\textcolor{blue}{28.90}} \tabularnewline
& VMI-FGSM~\cite{wang2021enhancing}
& {\textcolor{blue}{78.20}} & {\textcolor{blue}{83.20}} & 35.70 & 7.20 & 80.80 & 82.90 & 38.70 & 9.70 \tabularnewline

\hline
Ours & \emph{\Approach{}} & \textbf{94.10} & \textbf{93.70} & \textbf{86.50} & \textbf{34.50} & \textbf{94.10} & \textbf{93.70} & \textbf{86.50} & \textbf{34.50} \tabularnewline
\hline
\hline
 &  & 15.90 $\uparrow$ & 10.50 $\uparrow$ & 31.60 $\uparrow$ & 14.10 $\uparrow$ & 2.10 $\uparrow$ & 2.90 $\uparrow$ & 28.40 $\uparrow$ & 5.60 $\uparrow$ \tabularnewline
\hline
\end{tabular}\tabularnewline
\end{tabular}}}
\par\end{centering}
\caption{ASR results of black-box impersonation attack over CelebA-HQ dataset. Two offline models and two online commercial FR systems (Face++ and Microsoft) are used to evaluate attacking transferability. Our method uses IR152 FR and IR152 AR for white-box training, while other methods for comparisons are trained using two different FR models. The best-attacking performance results are shown in bold. The $2^{nd}$ place performance is shown in blue. The last row shows the promotion between best results vs. $2^{nd}$ results. 
\label{Tab:Overall for CelebA-HQ}}
\end{table*}

\begin{table*}[t!]
\begin{centering}
\setlength{\tabcolsep}{1.6mm}{
\hspace{-0.1in}\scalebox{0.9}{ %
\begin{tabular}{c}
\begin{tabular}{c|c|cc|cc|cc|cc}
\hline 
\multirow{4}{*}{Methods } & Dataset & \multicolumn{8}{c}{LFW} \tabularnewline
\cline{2-10}
& Source Model & \multicolumn{4}{c|}{IR152+FaceNet} & \multicolumn{4}{c}{IR152+IRSE50} \tabularnewline
\cline{2-10}
& \multirow{2}{*}{Target Model} & \multicolumn{2}{c|}{Offline Model} & \multicolumn{2}{c|}{Online Model} & \multicolumn{2}{c|}{Offline Model} & \multicolumn{2}{c}{Online Model} \tabularnewline
\cline{3-10}
& & IR50 & ResNet101 & Face++ & Microsoft & IR50 & ResNet101 & Face++ & Microsoft \tabularnewline
\hline
\multirow{6}{*}{Face-based} & Adv-Hat~\cite{komkov2019advhat}
& 1.80 & 9.30 & 1.80 & 0.10 & 5.00 & 13.40 & 2.20 & 0.10 \tabularnewline
& Adv-Glasses~\cite{sharif2016accessorize}
& 0.80 & 5.00 & 3.70 & 0.00 & 1.90 & 4.90 & 4.70 & 0.00 \tabularnewline
& Adv-Face~\cite{deb2020advfaces}
& 13.80 & 29.70 & 30.70 & 0.40 & 13.80 & 24.80 & 19.00 & 0.40 \tabularnewline
& Adv-Makeup~\cite{yin2021adv}
& 2.40 & 9.20 & 5.30 & 0.20 & 4.70 & 12.60 & 5.50 & 0.30 \tabularnewline
 & GenAP~\cite{xiao2021improving}
 & 4.20 & 13.60 & 15.20 & 0.30 & 4.30 & 14.50 & 13.90 & 0.50 \tabularnewline
\hline 
\multirow{4}{*}{Transfer-based} & PGD~\cite{madry2017towards}
& 75.80 & 78.20 & 46.70 & 19.10 & 89.30 & \textcolor{blue}{89.70} & 60.40 & 36.50 \tabularnewline
& TAP~\cite{zhou2018transferable}
& {\textcolor{blue}{76.90}} & {\textcolor{blue}{81.00}} & {\textcolor{blue}{54.10}} & \textcolor{blue}{28.60} & 89.60 & 89.60 & \textcolor{blue}{64.30} & \textcolor{blue}{45.60} \tabularnewline
& MI-FGSM~\cite{dong2018boosting}
& 68.40 & 71.00 & 41.90 & 21.10 & \textcolor{blue}{92.20} & 86.30 & 60.10 & 38.80 \tabularnewline
& VMI-FGSM~\cite{wang2021enhancing}
& {76.80} & 80.80 & 41.50 & 10.90 & 76.40 & 79.30 & 40.80 & 11.90 \tabularnewline
\hline
Ours & \emph{\Approach{}} & \textbf{98.70} & \textbf{98.60} & \textbf{96.10} & \textbf{59.30} & \textbf{98.70} & \textbf{98.60} & \textbf{96.10} & \textbf{59.30} \tabularnewline
\hline
\hline
 &  & 21.80 $\uparrow$ & 17.60 $\uparrow$ & 42.00 $\uparrow$ & 30.70 $\uparrow$ & 6.50 $\uparrow$ & 8.90 $\uparrow$ & 31.80 $\uparrow$ & 13.70 $\uparrow$ \tabularnewline
\hline
\end{tabular}\tabularnewline
\end{tabular}}}
\par\end{centering}
\caption{ASR results of black-box impersonation attack over LFW dataset. The settings are following Tab.~\ref{Tab:Overall for CelebA-HQ}.
\label{Tab:Overall for LFW}}
\end{table*}

\subsection{Why Select the AR Task?}

Theoretically, we have presented the high correlations between FR and AR in the earlier sections. Furthermore, we empirically explore the effectiveness of the AR task by quantitative analysis. Firstly, we compare the transferable ASRs results against various face-related task combinations in Tab.~\ref{Tab:FR Related Tasks}, where FR denotes the \emph{Face Recognition} task,  FLD indicates the \emph{Face Landmark Detection} task, FP means the \emph{Face Parsing} task, and AR denotes the \emph{Attribute Recognition} task. And then, all the combinations follow the basic Hard Parameter Sharing architecture to construct the joint-task attacking framework. Their attacking losses are in the same form as $\mathcal{L}_{adv}^{*}$ in Eq.~\ref{eq: cos_simi}. The ASR results demonstrate that the FR+AR outperforms all the competitors, which quantitatively proves that leveraging the AR task as a sibling task can craft more effective adversarial attacks. And the under-performance of FR+FLD and FR+FP combinations also indicate that not all face-related tasks can contribute to FR attacking transferability. In turn, it confirms the necessity of selecting appropriate face-related tasks as the attacking candidates for the FR task. 

\subsection{Experimental Results}
\label{sec:exp_res}

\noindent \textbf{Comparison with face-based methods.}
From Tab.~\ref{Tab:Overall for CelebA-HQ} and~\ref{Tab:Overall for LFW}, we observe that the patch-based methods have weak transferability on most target models as they are designed and tuned for physical attacks with small attacking areas. The results show that the adversarial examples attacking the entire face, i.e., Adv-Face, have the best transferability compared to all the other face-based methods. 
However, \emph{\Approach{}} can still significantly outperform Adv-Face.

\noindent \textbf{Comparison with transfer-based methods.}
We then compare \emph{\Approach{}} with four transfer-based attack methods (designed to generate strongly transferable adversarial examples). 
As observed from the results of CelebA-HQ in Tab.~\ref{Tab:Overall for CelebA-HQ}, \emph{\Approach{}} dominates all the transfer-based methods across various settings and evaluated models. Specifically, under the setting that uses IR152+FaceNet as white-box models, \emph{\Approach{}} outperforms the best results of transfer-based methods under offline models by $15.90\%$ on IR50 and $10.50\%$ on ResNet101. Meanwhile, \emph{\Approach{}} outperforms the best results of other competitors under online models by $31.60\%$ on Face++ and $14.10\%$ on Microsoft. Similarly, Tab.~\ref{Tab:Overall for CelebA-HQ} and~\ref{Tab:Overall for LFW} also show our superior performance. On average, \emph{\Approach{}} improves the state-of-the-art ASRs by 12.61\% and 55.77\% for offline pre-trained and online commercial models.

\begin{table}[t!]
\begin{centering}
\setlength{\tabcolsep}{1.0mm}{
\hspace{-0.1in}\scalebox{0.7}{ %
\begin{tabular}{c}
\begin{tabular}{c|ccc|cc|cc}
\hline 
\multirow{3}{*}{Methods } & \multicolumn{3}{c|}{Dataset} & \multicolumn{4}{c}{LFW}\tabularnewline
\cline{2-8}
& \multicolumn{3}{c|}{Source Model} & \multicolumn{2}{c|}{Offline Model} & \multicolumn{2}{c}{Online Model} \tabularnewline
\cline{2-8}
& IR152 & FaceNet & IRSE50 & IR50 & ResNet101 & Face++ & Microsoft \tabularnewline
\hline
\multirow{3}{*}{Single Model} & $\checkmark$ & - & - & 76.50 & 79.30 & 43.40 & 13.10\tabularnewline
& - & $\checkmark$ & - & 1.30 & 5.10 & 4.90 & 0.20\tabularnewline
& - & - & $\checkmark$ & 63.40 & 76.80 & 56.50 & 14.20\tabularnewline
\hline
\multirow{3}{*}{Ensemble}
& $\checkmark$ & $\checkmark$ & - & 75.80 & 78.20 & 46.70 & 19.10 \tabularnewline
& $\checkmark$ & - & $\checkmark$ & \textcolor{blue}{89.30} & \textcolor{blue}{89.70} & \textcolor{blue}{60.40} & \textcolor{blue}{36.50} \tabularnewline
& - & $\checkmark$ & $\checkmark$ & 65.80 & 77.90 & 59.20 & 16.80 \tabularnewline
\hline
\hline
\multirow{4}{*}{Ours} 
& 
\multicolumn{3}{c|}{Basic framework} & 80.90 & 92.20 & 69.80 & 37.20 \tabularnewline
& 
\multicolumn{3}{c|}{+ Hard P.S.} & 97.60 & 96.80 & 77.40 & 45.40 \tabularnewline
& 
\multicolumn{3}{c|}{+ JTMO} & 98.30 & 98.40 & 95.50 & 51.20 \tabularnewline
& 
\multicolumn{3}{c|}{+ CTGS} & \textbf{98.70} & \textbf{98.60} & \textbf{96.10} & \textbf{59.30}
\tabularnewline
\hline
\hline
\end{tabular}\tabularnewline
\end{tabular}}}
\par\end{centering}
\caption{Comparisons of ASR results of impersonation attack over LFW dataset. The ensemble represents the ensemble-training-based method. The $2^{nd}$ place results are shown in blue. 
\label{Tab:ablation_study_1}}
\end{table}

\subsection{Ablation Study}
\label{sec:ablation_study}
We study the impact of the different components of \emph{\Approach{}}. Specifically, \emph{\Approach{}} consists of the following components: (a) Hard Parameter Sharing (denoted as ``Hard P.S.''), (b) Joint-Task Meta Optimization (JTMO), and (c) Cross-Task Gradient Stabilizing (CTGS). As shown in Tab.~\ref{Tab:ablation_study_1}, we investigate the performance of \emph{\Approach{}} with different incorporated components. The line, ``\emph{Basic framework}'', is to directly average the adversarial losses for FR and AR models without optimization strategies. Its ASRs are competitive under single/ensemble model (single-task) settings. Empirically, this experimental result strongly supports the effectiveness of our idea that using the information from AR tasks can help boost attacking transferability against the FR task. 
Besides, we can observe that the ASRs gradually increase with adding each proposed component and significantly outperform other single/ensemble training-based competitors. Specifically, for two online models, Hard P.S. architecture boosts ASRs by $7.60\%$ and $8.20\%$, 
JTMO improves $18.10\%$ and $5.80\%$ compared with the Hard P.S. architecture. CTGS further achieves improvements of $0.60\%$ and $8.10\%$ compared with the Hard P.S. with JTMO. The results demonstrate the effectiveness of each proposed component in \emph{\Approach{}}.

\begin{figure}[t!]
    \centering
    \includegraphics[width=0.48\textwidth]{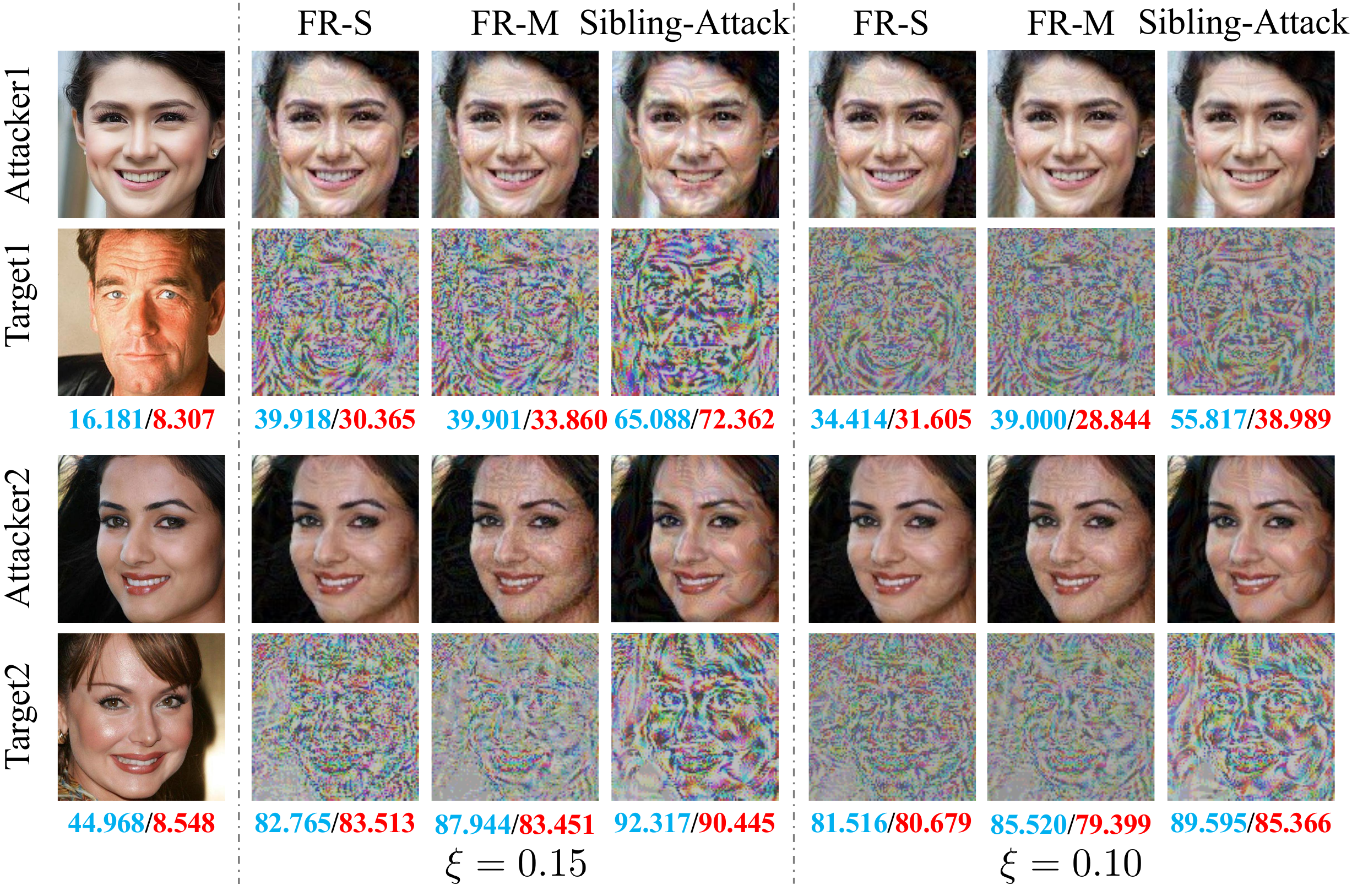}
    \centering
    \caption{Visualization of our adversarial perturbations comparing with attacks only against FR models. Each column shows the adversarial example and its post-processed perturbations. Query results from Face++ and Microsoft are shown in blue and red.} 
    \label{fig:visual}
\end{figure}

\subsection{Visualization and Analysis}
\noindent\textbf{Visualization of adversarial perturbations.}
In Fig.~\ref{fig:visual}, we visualize the generated adversarial examples/perturbations from FR-S (against a single FR model, IR152), FR-M (against two ensemble FR models, IR152 and FaceNet), and \emph{\Approach{}} for two pairs of target and attacker images from CelebA-HQ. For fair comparisons, we select a cross-gender and a same-gender pair. The first column presents the legitimate examples with query results, and the following columns present the adversarial examples/perturbations under two different $L_\infty$ bounds ($\xi$=0.10,0.15) with the query results from Face++ and Microsoft, respectively. Specifically, we post-process the perturbations to make them more perceptible. In detail, we multiply all the perturbation values by $5$ then truncate the values less than $\xi/3$, then project each perturbation value into $[0, 255]$ for better visualization. We can discern a salient shape of a face and some facial components in the adversarial perturbations generated by \emph{\Approach{}}, which are different from the perturbations generated by FR-S and FR.

\begin{figure}[t!]
    \centering
    \includegraphics[width=0.48\textwidth]{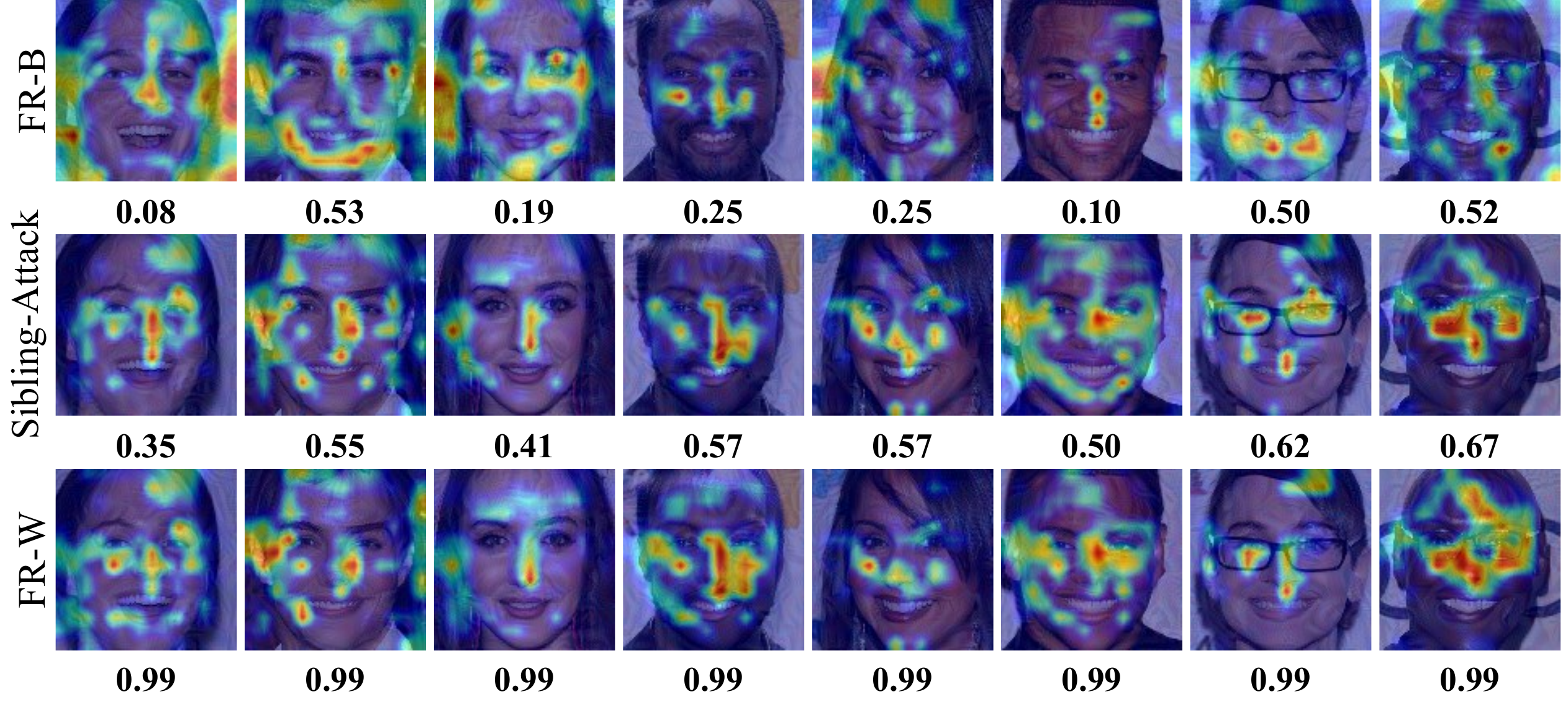}
    \centering
    \caption{Visualization using Grad-CAM~\cite{selvaraju2017grad} produces attention maps on an offline FR model (IRSE50). We display the similarity score between the attacker and the target face on the FR model under each picture. Best viewed in color.}
    \label{fig:vis}
\end{figure}

\noindent\textbf{Visualization of black-box adversarial gradient responses.} We further explore why adversarial examples generated by \emph{\Approach{}} exhibit more attacking transferability by employing Grad-CAM~\cite{selvaraju2017grad}, as shown in Fig.~\ref{fig:vis}. For each row, we visualize the gradient responses. Specifically, FR-B (Black) denotes the black-box scenario, ensemble attacking IR152 and FaceNet using PGD and visualizing Grad-CAM on IRSE50. FR-W (White) denotes the white-box scenario, ensemble attacking IRSE50 via PGD and visualizing Grad-CAM on IRSE50. Notably, the gradient responses in FR-W serve as the ground truth for measuring the attacking transferability of each approach. Specifically, the more visual similarity in gradient responses between the evaluated approach and ground truth implies stronger transferability. We can observe that gradient responses in FR-B seem either (1) to pay more attention to the background or (2) overfit to some local facial regions. In contrast, gradient responses from \emph{\Approach{}} and the target model both focus more on the similar key facial regions, which interprets the stronger transferability of \emph{\Approach{}}.

\begin{figure}[t!]
    \centering
    \includegraphics[width=0.47\textwidth]{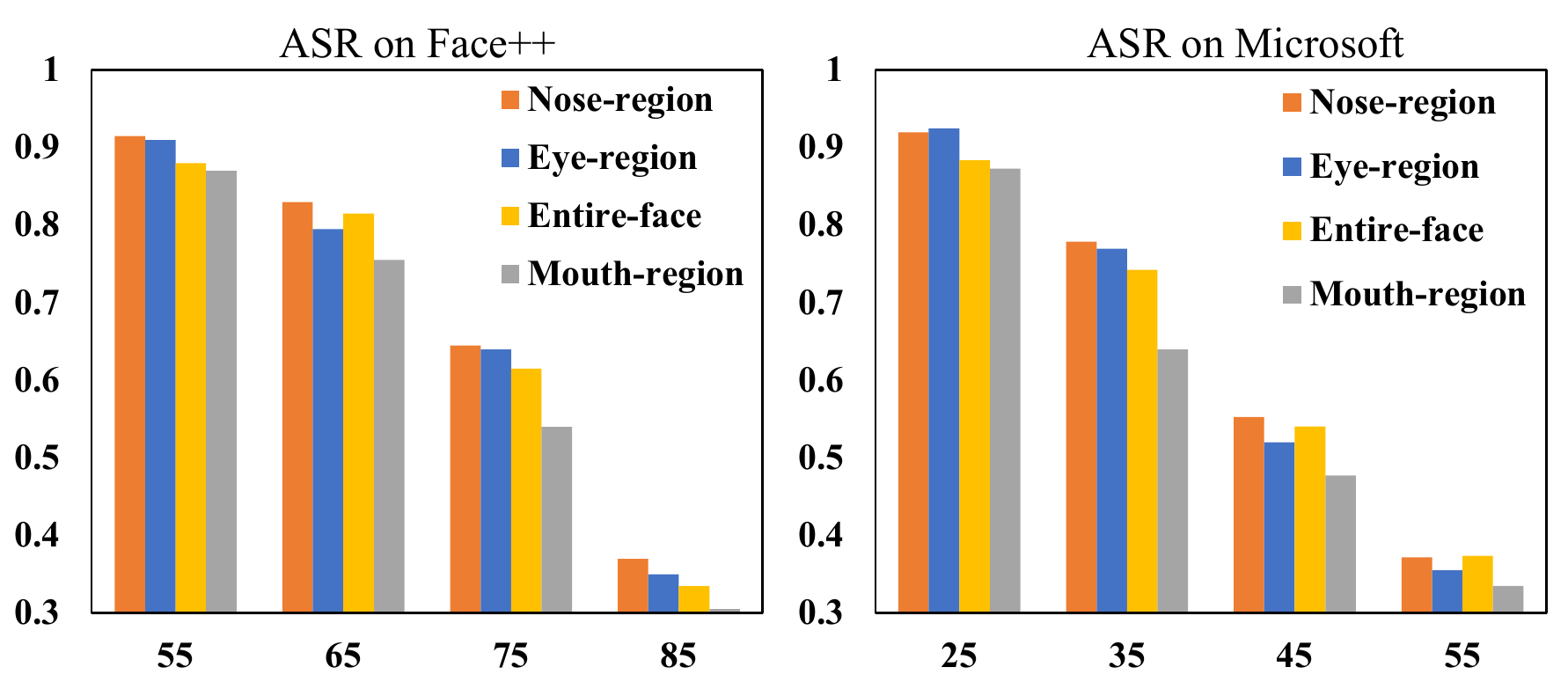}   
    \centering
    \caption{ASR on Face++ and Microsoft of \emph{\Approach{}} with AR models trained by different facial attribute groups. The x-axis represents the similarity score. The y-axis represents the ASR under the corresponding similarity score. }
    \label{fig:AR_different_transfer}
\end{figure}

\noindent\textbf{Visually-indistinguishable analysis.}
In addition to the efficacy, we also analyze the visual indistinguishability of our crafted adversarial samples. We use \textit{Structural Similarity} (SSIM) ~\cite{wang2002universal} and the \textit{Mean Square Error} (MSE)~\cite{marmolin1986subjective} between basic examples and corresponding adversarial examples as metrics. As shown in Table.~\ref{Tab:ssim_mse}, we compare SSIM and MSE with the other invisible attacking methods on LFW. The results demonstrate that the SSIM and MSE of \textit{\Approach{}} are competitive with other methods. Last but not least, \textit{\Approach{}} can achieve a much better-attacking transferability against black-box FR models. 

\noindent\textbf{Transferability analysis of different facial attributes.}
Our experimental results for \emph{\Approach{}} have already shown that exploiting an auxiliary AR model can help generate strongly transferable adversarial examples against FR models. This section further explores which facial attributes can bring more transferability to the FR task. Specifically, we divide the 18 facial attributes used for training our IR152 AR model into four non-overlapping groups by position (eye-region, nose-region, mouth-region, other-region). As Fig.~\ref{fig:AR_different_transfer} shows, mouth-region facial attributes bring weaker transferability to the FR task than attributes in other regions, which is consistent with some
existing works~\cite{diniz2020face,xiao2021improving} on face recognition and face-based attacks.

\begin{table}[t!]
\renewcommand\arraystretch{1.1}
\begin{centering}
\setlength{\tabcolsep}{1.6mm}{
\hspace{-0.1in}\resizebox{0.45\textwidth}{!}{ %
\begin{tabular}{c}
\begin{tabular}{c|cc|cc}
\hline 
 Dataset & \multicolumn{4}{c}{LFW} \tabularnewline
\cline{1-5}
 Source Model & \multicolumn{2}{c|}{IR152+FaceNet} & \multicolumn{2}{c}{IR152+IRSE50} \tabularnewline
\cline{1-5}
 Metrics & SSIM & MSE & SSIM & MSE \tabularnewline
\hline
PGD~\cite{madry2017towards} & {\color{blue}0.619} & \textbf{175.915} & {\color{blue}0.594} & {\color{blue}193.801} \tabularnewline
 TAP~\cite{zhou2018transferable} & 0.613 & {\color{blue}181.279} & 0.591 & 196.942 \tabularnewline
 MI-FGSM~\cite{dong2018boosting} & 0.473 & 343.227 & 0.463 & 350.162 \tabularnewline
 VMI-FGSM~\cite{wang2021enhancing} & 0.588 & 200.418 & 0.574 & 215.346 \tabularnewline
 \hline
\emph{\Approach{}} & \textbf{0.626} & 187.491 & \textbf{0.626} & \textbf{187.491} \tabularnewline
\hline
\end{tabular}\tabularnewline
\end{tabular}}}
\par\end{centering}
\caption{SSIM and MSE scores of our methods and other competitors. The best results are shown in bold. The $2^{nd}$ place performance is shown in blue.
\label{Tab:ssim_mse}}
\end{table}

\begin{table}[t!]
\begin{centering}
\setlength{\tabcolsep}{1.6mm}{
\hspace{-0.1in}\scalebox{0.79}{ %
\begin{tabular}{c}
\begin{tabular}{c|cccc}
\hline 
Group & Eye-region & Nose-region & Mouth-region & Other-region \tabularnewline
\hline
Baseline AR & 148.85 & 223.97 & 184.77 & 201.79 \tabularnewline
Ours & \textbf{162.41} & \textbf{241.29} & \textbf{195.46} & \textbf{214.02} \tabularnewline
\hline
\end{tabular}\tabularnewline
\end{tabular}}}
\par\end{centering}
\caption{Comparisons of the overall prediction difference between \emph{\Approach{}} and a baseline white-box AR for four attribute groups after attacking a black-box AR model. 
\label{tab:AR_transfer}}
\end{table}

\noindent\textbf{Transferability analysis against AR tasks.} We also explore whether attackers can adopt adversarial information from the FR model to improve the transferability against the black-box target AR model by \emph{\Approach{}}. As shown in Tab.~\ref{tab:AR_transfer}, we compare the overall attributes prediction difference across $1000$ image pairs for four attribute groups. The overall attributes prediction changes can be computed as:
\begin{equation}
\text{Overall Pred. Diff.} = \sum_{s}^{S} \| \mathcal{A}_{B}(x_{adv}^s)-\mathcal{A}_{B}(x^s) \|_1
\label{eq: overall diff}
\end{equation}
where $ \mathcal{A}_{B}(\cdot) $ denotes the target AR model, which outputs the predicting score for each attribute, and $S = 1000$, $ x_{adv}^s$ is the adversarial example crafted by attacking our model or the baseline white-box AR model. The results indicate that \emph{\Approach{}} can conduct more prediction difference than the competitor, which supports that our method can also boost the attacking transferability against the AR model. 


\section{Conclusion.} 
The proposed \emph{\Approach{}} firstly leverages a highly FR-related task AR as the sibling task to generate strongly transferable adversarial attacks against FR tasks under the black-box setting. It mainly focuses on digital scenarios, but it is equally essential for face recognition security as to the physical attacks since it can reveal more threatening adversarial risks. Besides, the proposed method may be used maliciously to hazard the security of existing FR models in real life, the adversarial training and de-noise strategies can mitigate the negative impacts. Extensive experiments demonstrate the superior transferability of \emph{\Approach{}} on various offline and online commercial FR models. In the future, we also intend to extend the proposed idea to other computer vision and biometrics tasks besides FR.

\section*{Acknowledgments}
This work was supported by NSF CNS 2135625, CPS 2038727, CNS Career 1750263, and a Darpa Shell grant. 

{\small
\bibliographystyle{IEEE_fullname}
\bibliography{cvpr2022}
}

\clearpage

\end{document}